\documentclass{article}

\usepackage{PRIMEarxiv}

\usepackage[utf8]{inputenc} 
\usepackage[T1]{fontenc}    
\usepackage[hidelinks]{hyperref}      
\usepackage{url}            
\usepackage{booktabs}       
\usepackage{amsfonts}       
\usepackage{nicefrac}       
\usepackage{microtype}      
\usepackage{lipsum}
\usepackage{graphicx}
\graphicspath{{media/}}     

\usepackage{algorithm,algpseudocode}
\usepackage{acro}
\usepackage{graphicx}
\usepackage{float}
\usepackage{multirow}
\usepackage{enumitem}
\usepackage{pgfplots}
\usepackage{pgfplotstable}
\usepackage{amsfonts}
\usepackage{amsmath,amstext}
\usepackage{textcomp}
\usepackage{comment}
\usepackage{gensymb}
\usepackage{balance}
\pgfplotsset{compat=newest}
\usepgfplotslibrary{groupplots,dateplot}
\usepackage{pgfplotstable}
\usepgfplotslibrary{statistics}
\usepgfplotslibrary{patchplots}
\usepackage[capitalise]{cleveref}
\usetikzlibrary{pgfplots.statistics, pgfplots.colorbrewer,pgfplots.dateplot,
        shadows,
        matrix} 
\usepackage{multicol}
\usepackage{algorithm}
\usepackage{algpseudocode}
\usepackage{wrapfig}
\usepackage[font=small,labelfont=bf]{caption}
\usepackage{enumitem,kantlipsum}
\usepackage[round]{natbib}

\usepackage{balance} 
\usepackage{booktabs}

\DeclareMathOperator*{\argmin}{arg\,min}
\title{Act as You Learn: Adaptive Decision-Making in Non-Stationary Markov Decision Processes\thanks{Recommended Citation: Luo, Baiting, Zhang, Yunuo, Dubey, Abhishek, and Mukhopadhyay, Ayan. ``Act as You Learn: Adaptive Decision-Making in Non-Stationary Markov Decision Processes.'' \textit{International Conference on Autonomous Agents and MultiAgent Systems (AAMAS)}. 2024.}}

\author{
  Baiting Luo, Yunuo Zhang, Abhishek Dubey, Ayan Mukhopadhyay\\
  Institute for Software Integrated Systems\\
  Vanderbilt University \\
  Nashville, TN 37212, USA\\
  \texttt{\{baiting.luo, yunuo.zhang, abhishek.dubey, ayan.mukhopadhyay\}@vanderbilt.edu} \\
}

\newcommand{\BibTeX}{\rm B\kern-.05em{\sc i\kern-.025em b}\kern-.08em\TeX}

\begin{document}
\maketitle

\begin{abstract}
A fundamental (and largely open) challenge in sequential decision-making is dealing with non-stationary environments, where exogenous environmental conditions change over time. Such problems are traditionally modeled as non-stationary Markov decision processes (NSMDP). However, existing approaches for decision-making in NSMDPs have two major shortcomings: first, they assume that the updated environmental dynamics at the current time are known (although future dynamics can change); and second, planning is largely pessimistic, i.e., the agent acts ``safely'' to account for the non-stationary evolution of the environment. We argue that both these assumptions are invalid in practice---updated environmental conditions are rarely known, and as the agent interacts with the environment, it can learn about the updated dynamics and avoid being pessimistic, at least in states whose dynamics it is \textit{confident} about. We present a heuristic search algorithm called \textit{Adaptive Monte Carlo Tree Search (ADA-MCTS)} that addresses these challenges. We show that the agent can learn the updated dynamics of the environment over time and then \textit{act as it learns}, i.e., if the agent is in a region of the state space about which it has updated knowledge, it can avoid being pessimistic. To quantify ``updated knowledge,'' we disintegrate the \textit{aleatoric} and \textit{epistemic} uncertainty in the agent's updated belief and show how the agent can use these estimates for decision-making. We compare the proposed approach with the multiple state-of-the-art approaches in decision-making across multiple well-established open-source problems and empirically show that our approach is faster and highly adaptive without sacrificing safety. 
\end{abstract}

\pagestyle{fancy}
\fancyhead{}


\maketitle

\section{Introduction}

Sequential decision-making under uncertainty is present in many real-world problems, such as vehicle routing, emergency management, medical diagnosis, and autonomous driving~\citep{aradi2020survey,mukhopadhyay2022review,DBLP:journals/corr/abs-2307-11876,DBLP:conf/iccps/LuoRPKKM23}. A fundamental problem in such domains is to model and optimize the ability of an agent to adapt to changing environmental dynamics~\citep{satia1973markovian,white1994markov,DBLP:conf/aspdac/0002HJLLLWWX21,DBLP:conf/icml/0001ZJWJYW0023}. For example, consider the dispatch of emergency responders in a city; once a policy is learned, a first responder agency can optimally dispatch resources. However, the city's traffic conditions can change, rendering the learned policy stale. Our goal is to systematically model, investigate, and optimize how agents can adapt to such changing environments.
Our problem setting is related to prior work on Lifelong Reinforcement Learning (LRL)~\citep{thrun1995lifelong}, where an agent's tasks (or environments) evolve over time. Typically, in such settings, agents can \textit{transfer} knowledge from previous tasks, explore the new environment, and learn a new policy for the updated environment. We look at a somewhat related but mostly orthogonal problem---learning the new policy takes time, and \textit{as the agent learns the new policy, how can it keep making decisions?} In the example of emergency response, the agency can learn a new optimal policy for the updated road conditions, but it must dispatch responders (near optimally) to incidents while the policy is learned.

A principled model to use in our problem setting is a non-stationary Markov decision process (NS-MDP), proposed by \citeauthor{DBLP:conf/nips/LecarpentierR19}~\citeyearpar{DBLP:conf/nips/LecarpentierR19}. Intuitively, an NS-MDP can be thought of as an MDP whose state space contains an identifier for time. If the task is episodic and the agent can repeatedly interact with the environment, the addition of time in the state space is trivial to deal with; however, if the interactions are limited (as in decision-making in real-world deployed applications), dealing with the updated environment is challenging~\citep{DBLP:conf/nips/LecarpentierR19}. Naturally, if the MDP changes rapidly and arbitrarily, all hope is lost. One simplifying assumption is to restrict the rate of change, e.g., enforce that the evolution of the MDP (i.e., its transition and reward function) follows Lipschitz Continuity. Intuitively, this assumption means that the changes occur slowly (albeit continuously) over time. At any time step, it is typically assumed that the agent \textit{knows the current MDP} but does not know the future evolution and, therefore, must plan under uncertainty and make decisions~\citep{DBLP:conf/nips/LecarpentierR19}. 

The state-of-the-art online approach in these settings is to use risk-averse tree search~\citep{DBLP:conf/nips/LecarpentierR19}, i.e., the agent accounts for the uncertain future evolution of the environmental dynamics by ``playing safe.'' While this approach can work well in some situations, it treats the lack of knowledge about the non-stationary evolution as a monolithic concept. We challenge and tackle two assumptions in such approaches: \underline{first}, we argue that more often than not, the current environmental conditions are not known to the agent~\citep{DBLP:conf/iclr/AntonoglouSOHS22}, especially in complex real-world problems. Indeed, even in online settings, the agent must interact with the environment to learn and adapt to the updated dynamics. \underline{Second}, we argue that the lack of knowledge about the non-stationarity is not monolithic---as the agent interacts with the environment, there are regions of the state space that it learns more about than other regions. During decision-making, the agent can avoid ``playing safe'' in the known regions, achieving significantly improved outcomes.\footnote{Admittedly, we use \textit{lack of knowledge} rather loosely here; we clarify how to quantify this idea later.}

We begin by briefly introducing our problem setting. We consider an agent that must make sequential decisions under uncertainty. At some discrete time point, the environment, i.e., the transition function of the MDP, changes. While the agent does not know of \textit{what} the change is, we assume that the agent knows that a change has occurred. In practice, the agent can be notified of a change by data-driven anomaly detectors. Notified of a change but unaware of what the change is, the agent must adapt to the new environment. We propose an online approach that explores the environment safely to gather data and then estimates the updated transition dynamics. The learned dynamics can have both epistemic (due to the lack of data collected from the new environment) and aleatoric uncertainty (due to the inherent stochasticity of the environment). Our approach leverages these uncertainties and behaves in a pessimistic (i.e., risk-averse) manner in regions of the state space that it is unfamiliar with while being more risk-seeking in areas of the state space that it has adapted to. Specifically, we make the following contributions:

\begin{enumerate}
    \item We show how an agent can start with pessimistic decision-making to explore a new environment safely in a non-stationary MDP and use a Bayesian learning approach to refine its estimates of the updated environmental parameters.
    \item We build upon risk-averse minimax search and propose Adaptive Monte Carlo Tree Search (ADA-MCTS), an online heuristic search algorithm that adapts standard Monte Carlo tree search to our problem setting.
    \item We show how NS-MCTS can balance performance and risk by employing a dual-phase adaptive sampling strategy. As the agent spends more time in the environment, it adapts more, thereby reducing its pessimism and focusing more on performance.
    \item We evaluate our approach on three well-established benchmark settings and show that our approach significantly outperforms baseline approaches.
    \item We perform an ablation study to evaluate the importance of each component of the proposed approach.
\end{enumerate}

\section{Problem}

We begin by providing a description of non-stationary MDPs in our context and then outline the exact goals that we seek to achieve
. 
\subsection{Non-Stationary Markov Decision Process (NS-MDP)}

An MDP, denoted as $M$, is characterized by the tuple $(\mathcal{S}, \mathcal{A}, p, r, \gamma)$, where $\mathcal{S}$ represents the set of possible states, $\mathcal{A}$ denotes the set of actions, $p(s' \mid s, a)$ is the transition function that denotes the probability of reaching state $s'$ if action $a$ is taken in state $s$, $r(s, a)$ represents the reward function that quantifies the utility of taking action $a$ in state $s$, and $\gamma$ is the discount factor. The agent's goal is to maximize expected discounted additive rewards over a pre-specified time horizon; here, we consider the infinite horizon case. In offline settings, the goal of the agent is to learn an optimal policy, which is a mapping from states to actions. In the online setting, the agent's goal is to compute a near-optimal action for the current state; the agent takes this action and transitions to a new state, where the online planner is invoked again. 

We look at a non-stationary setting, where the MDP evolves over time. Unlike prior work by \citeauthor{DBLP:conf/nips/LecarpentierR19}~\citeyearpar{DBLP:conf/nips/LecarpentierR19}, we assume that the MDP changes at discrete points in time. We hypothesize that discrete changes model the real-world applications more closely; e.g., consider traffic in a city---while traffic evolves continuously, such variations are usually subsumed within a given stochastic process model, i.e., the original MDP. Sudden discrete changes, on the other hand, might not be subsumed in the original model. 
Specifically, we denote the $k-1$-th version of an MDP at an arbitrary time step as $M_{k-1}$. At some point in the future, the MDP evolves to $M_{k}$, e.g., the transition function of the MDP changes. The key difference between our setting and that of lifelong reinforcement learning is the very nature of the goals---while the latter seeks to learn an optimal policy given the new MDP, we focus on online planning instead. Note that these two goals are not contradictory; rather, they are actually complementary. We prescribe that the agent \textit{should} learn a new optimal policy given that the system dynamics have changed. However, while it learns the new policy (which is typically done offline), the agent must still make decisions in the real world, which can be facilitated through the proposed online planning approach. 

Our problem setting can be summarized as follows: at an arbitrary time $t$, the agent's decision-making problem is summarized by the MDP $M_1$. While the agent might not know the exact transition function $p_1$ for $M_1$, we assume that it has access to a (potentially imperfect) model $\hat{p}_1$. Now, some time elapses, and at time $t'$ (say), the agent finds itself in MDP $M_2$, with an updated transition function ${p}_2$. Naturally, the agent does not know ${p}_2$. The agent's goal is to make decisions online in $M_2$. As the agent makes decisions and interacts with the environment, it gathers data $D_2$. The agent can then use $D_2$ and $\hat{p}_1$ to estimate $\hat{p}_2$. Note that this process is repeated over time---as the agent gathers more data, its estimates of $\hat{p}_2$ improve, thereby enabling it to make better decisions. Given the problem formulation, we reiterate the differences between our problem setting and that proposed by \citeauthor{DBLP:conf/nips/LecarpentierR19}~\citeyearpar{DBLP:conf/nips/LecarpentierR19}: 1) while they assume that a perfect snapshot of the environment at time $t'$ is available to the agent, we challenge this assumption and claim that in practice, the agent must learn about the environment by interacting with the environment; 2) they assume that the environment evolves continuously, while we focus on discrete changes to the MDP; and 3) we hypothesize that as the agent interacts with the environment, it can avoid being pessimistic during planning, at least in some regions of the state space (unlike prior work).




\subsection{Goals}

Consider two MDPs, $M_{k-1}$ and $M_{k}$, where $M_{k}$ represents the agent's current decision-making model. Given our problem setting, we have two major goals: 1) we seek to design an algorithmic approach that can perform relatively safe exploration of the new environment. However, we want to avoid unnecessary pessimism in the agent's behavior, i.e., the agent must be able to exploit its newly-acquired knowledge for (near-optimal) planning. Therefore, \textit{our first goal is to design an approach that adeptly balances the goals of safe exploration and reward maximization}. 2) Initially, the agent lacks any explicit knowledge pertaining to the dynamics of $M_{k}$. As the agent collects data by interacting with the new environment, it must be able to update its existing knowledge. We hypothesize that if the environment has not changed drastically (otherwise, adaptation is pointless), the agent can assimilate and reconcile past knowledge encapsulated
by $M_{k-1}$ with the new data, to estimate the dynamics of $M_{k}$. Crucially, the agent must be able to identify how \textit{well} does its model of the new environment work. Therefore our second goal is \textit{to develop a principled approach for the agent to learn the dynamics of the updated environment (and estimate the quality of the learned model)}.

\section{Approach}

The problem of finding optimal actions in the new environment becomes trivial if we have direct access to an exact environment model or sufficient samples from $M_{k}$ beforehand. However, such assumptions are unrealistic as the future model is generally unknown. Given the dynamic nature of transitions, notably between $M_{k-1}$ and $M_{k}$, such dynamism potentially renders prior experience data obsolete due to the non-stationarity of the environment, and we often find ourselves devoid of data samples for impending transitions. A viable approach in these situations is to employ robust algorithms to collect representative samples from the evolving environment. Such models, in turn, become pivotal for adaptive decision-making, enabling principled decision-making based on the observed environmental stochasticity.

We begin by introducing risk-averse MCTS (RA-MCTS), which is the Monte Carlo Tree Search extension of the worst-case approach.

\subsection{Risk-Averse Monte Carlo Tree Search}
Tree search algorithms such as minimax search, alpha-beta pruning, and expectiminimax have significantly contributed to solving deterministic or stochastic decision-making problems in both single-agent and adversarial settings~\cite{kochenderfer2022algorithms}.~\cite{DBLP:conf/nips/LecarpentierR19} presented the risk-averse tree-search algorithm that extends standard tree search methods for stochastic control problems to operate in non-stationary environments. To consider the inherent stochasticity in uncertain environments,\footnote{note that here, we refer to the uncertainty in any stochastic control process and not the non-stationarity of NS-MDPs.} they consider trees with alternating \textit{decision} nodes and \textit{chance} nodes. The key idea behind their approach is elegant---since the future evolution of the MDP is non-stationary, it is prudent to plan under the worst possible evolution. This idea is realized through the minimax algorithm, where the max operator corresponds to the agent's policy, and the min operator captures the worst evaluation of the NS-MDP~\cite{DBLP:conf/nips/LecarpentierR19}.

A distinctive approach is Monte Carlo Tree Search (MCTS), which leverages Monte Carlo sampling to estimate the values of state-action pairs. A critical component of MCTS is using Upper Confidence Bounds for Trees (UCT), expressed as $\text{UCT}(s,a) = Q(s,a) + c \sqrt{\frac{\log(N(s))}{N(s,a)}}$,
where $Q(s,a)$ represents the estimated value of executing action $a$ in state $s$, $N(s)$ denotes the number of times state $s$ has been visited, $N(s,a)$ signifies the number of times action $a$ has been chosen in state $s$, and $c$ is an exploration coefficient. The UCT formula adeptly balances exploration and exploitation during the search process, promoting exploration of less-visited state-action pairs while also exploiting promising actions based on accumulated knowledge. 



Our first contribution lies in adapting standard MCTS to a risk-averse framework. We use a similar idea as~\cite{DBLP:conf/nips/LecarpentierR19}; however, our key challenge is adapting the minimax approach to UCT-based action selection. Consider an agent whose decision-making problem at the current time step $t'$ is modeled by the MDP $\hat{M_k}$. The agent does not know $M_k$ exactly; rather, it has an imperfect model of it, i.e., while it does not know the exact transition function $p_k$, it has access to an estimate $\hat{p}_k$. The crux of our approach lies in computing the state-action values by using a pessimistic view of $\hat{p_k}$ (this basic idea of the risk-averse search is inspired from prior work by~\cite{DBLP:conf/nips/LecarpentierR19}). For an arbitrary policy $\pi$, we denote pessimistic $V$ and $Q$ functions as follows (we drop the reference to $k$ and simply use $\hat{p}$ for readability):

\begin{align}
V^{\hat{\pi}}_t(s) & := \min_{\hat{p}} \mathbb{E} \left[ \sum_{t = t'}^{\infty} \gamma^{t'} R(s_t, a_t) \bigg\vert^{s_t = s}_{a_t \sim \pi(\cdot \mid s_t), \, s_{t+1} \sim \hat{p}(\cdot \mid s_t, a_t)} \right] \\
Q^{\hat{\pi}}(s,a) & := \min_{\hat{p}} \mathbb{E}_{s' \sim p_{wc}(\cdot | s, a; \hat{M}_k)} \left[R(s,a) + \gamma V^{\hat{\pi}}_{t+1}(s')\right]
\end{align}
The optimal policy counterparts of pessimistic decision-making can then be represented as:
\begin{align}
V^{*}(s) & := \max_{a \in \mathcal{A}} Q^{*}(s,a) \\
Q^{*}(s,a) & := \min_{\hat{p}} \mathbb{E}_{s' \sim \hat{p}} \left[R(s,a) + \gamma V^{*}_{t+1}(s')\right]
\end{align}
Given this setting, our goal is to compute these quantities within the MCTS algorithm. 

We use the same terminology as~\cite{DBLP:conf/nips/LecarpentierR19} to enable readers to cross-reference between risk-averse (minimax) tree search~\cite{DBLP:conf/nips/LecarpentierR19} and risk-averse MCTS. Let $\nu^{s}$ and  $\nu^{s,a}$ denote an arbitrary decision node and chance node within MCTS, respectively. To adapt the min-max nature of risk-averse tree search to MCTS, we use the \textit{min} operator in the chance nodes and the \textit{max} operators in the decision nodes. Specifically, 
\begin{align}
V(\nu^{s}) = \max_{a \in \mathcal{A}} \left[ V(\nu^{s,a}) + c \sqrt{\frac{\log(N(\nu^{s}))}{N(\nu^{s,a})}} \right]\\
V(\nu^{s,a}) = R(s,a) + \gamma \mathbb{E}_{s' \sim p_{wc}(\cdot | s, a)} V(\nu^{s'})
\end{align}
where $p_{wc}$ denotes a worst-case sampling strategy with respect to $p$. Specifically, $p_{wc}=(0,\cdots,0,1,0,\cdots,0)$, i.e., a vector where every element is $0$, except $\argmin_{i} \nu_{i}$, with $\nu_{i}$ denoting the value of the $i$th state reachable from the chance node.



In this formulation, the $\min$ operator in the chance nodes' function represents the pessimistic selection strategy, ensuring that the decision-making process is geared towards mitigating the risks associated with the worst-case scenarios. This approach is particularly beneficial in uncertain or adversarial environments where a more conservative strategy is desirable to avoid significant losses. The UCT exploration term $\sqrt{\frac{\log(N(\nu^{s}))}{N(\nu^{s,a})}}$ encourages exploration of less-visited state-action pairs, thus providing a balance between exploring the state-action space and adhering to a risk-averse decision-making framework.

\subsection{Dual Phase Adaptive Sampling}
\begin{figure}[t]
    \centering
    \includegraphics[width=0.5\columnwidth]{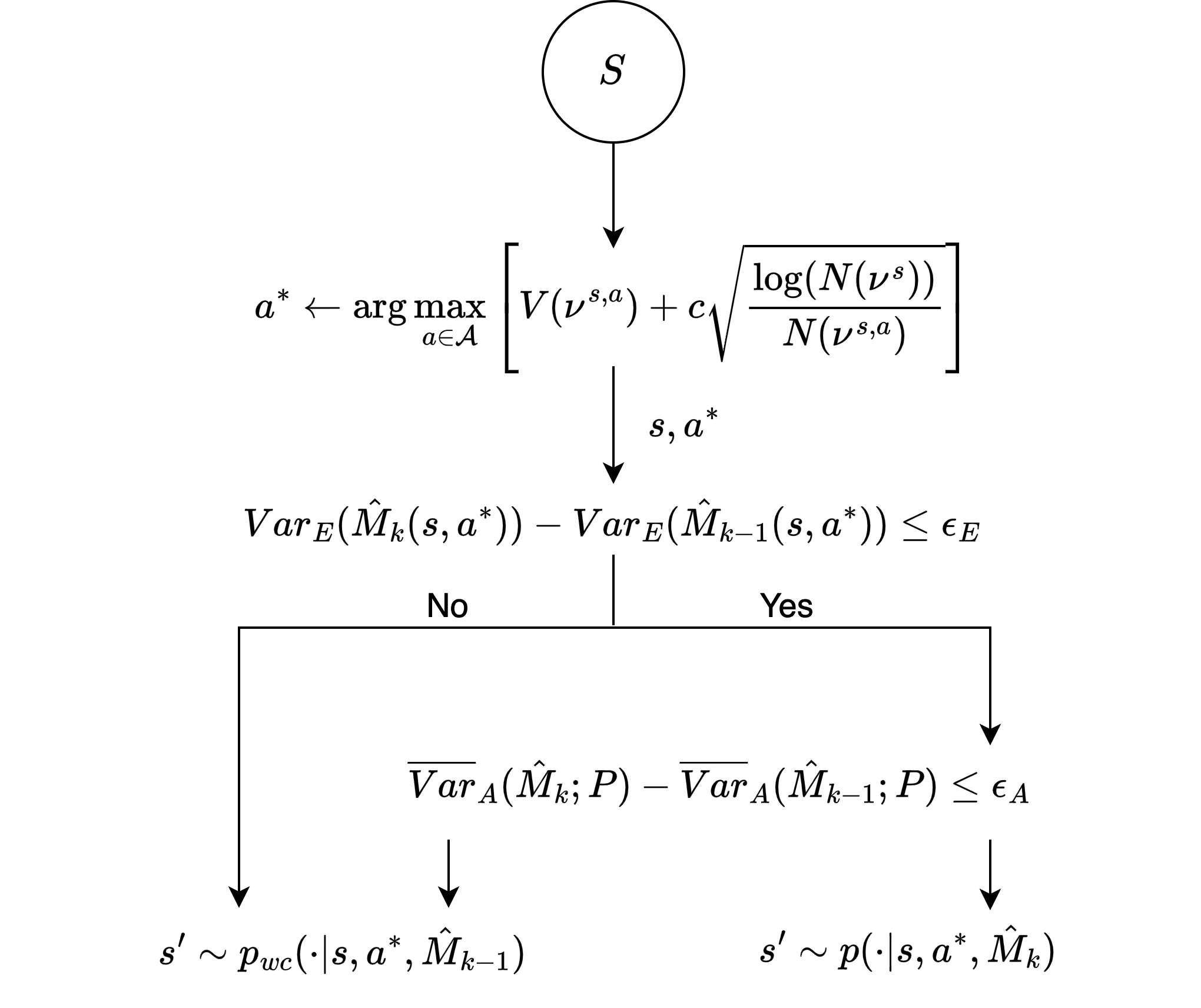}
    \caption{Schematic representation of the proposed dual-phase adaptive sampling approach.}
    \label{fig:sampling}
\end{figure}

Having defined risk-averse MCTS, we now address the challenge of adapting the agent's search strategy as it explores the new environment. Recall that our goal is to leverage the knowledge gathered through interacting with the new environment to refine our decision-making approach. We propose to guide this refinement by the observed level of stochasticity in the environment and the uncertainties associated with our model's understanding of the environment. Specifically, we use \textit{epistemic} uncertainty to gauge the model's familiarity with the state-action space of the new MDP and \textit{aleatoric} uncertainty to assess the inherent randomness of the environment. Our approach is based on the following idea---if the agent's epistemic uncertainty about a region of the state space is high, the agent must be risk-averse, gather more data, refine its understanding, and then proceed with potentially (relatively) risky actions that maximize rewards. Similarly, if the agent perceives the aleatoric uncertainty to be high, it must be risk-averse as it is difficult for it to compute stable estimates (i.e., low variance) of its expected utility.

We now describe how to leverage this idea within a search tree, where we must sample the next states based on a noisy and imperfect model of the environment. We refine the agent's sampling strategy based on epistemic and aleatoric uncertainty; the essence of this phase is the dynamic pivoting of the strategy in response to the environmental conditions and the model's understanding thereof. Initially, a conservative approach is adhered to, ensuring adequate exploration of the state space. Once a certain level of exploration is achieved, as indicated by a reduction in epistemic uncertainty, the sampling strategy transitions to from a risk-averse setting to a ``regular'' setting based on the environmental stochasticity assessed through aleatoric uncertainty (if the aleatoric uncertainty is high, the agent continues with a risk-averse sampling strategy). We present the dual-phase adaptive approach in Fig~\ref{fig:sampling}.
To quantify these uncertainties, a variety of methods, including bootstrapping~\citep{mooney1993bootstrapping}, variational inference~\citep{blei2017variational}, and Bayesian neural networks (BNNs)~\citep{lampinen2001bayesian}, could be potentially applied. BNNs, with their inherent probabilistic nature, effectively capture both epistemic and aleatoric uncertainties~\citep{DBLP:conf/nips/KendallG17}. However, our approach is versatile and not strictly bound to BNNs. As long as a model can accurately estimate both types of uncertainties, capture the dynamics of the environment, and ensure prompt inference, it can be integrated into our adaptive sampling framework.

\textbf{Bayesian Neural Networks for Uncertainty Quantification:} In our algorithm, we draw $N$ samples from the BNN's posterior weight distribution to obtain $N$ different sets of predictions, each characterized by a mean and a variance. Specifically, let the agent's environment be denoted by $M_{k}$. Then, given a state-action pair $(s, a)$, the output $\hat{p}_{k}$ (estimated transition function for $\hat{M}_{k}$), approximated by a BNN, is represented as:
\begin{equation}
\hat{p}_{k}(s,a) = (\mu(s,a; W), \sigma^2(s,a; W))
\end{equation}
where $\mu(s,a; W)$ and $\sigma^2(s,a; W)$ are respectively the predicted mean and the predicted variance of the output for state-action pair $(s, a)$ given neural network weights $W$. 
When quantifying uncertainties, the distinction between aleatoric and epistemic uncertainties often necessitates different approaches to estimation. For aleatoric uncertainty, which captures the inherent data noise or randomness, increasing the number of samples can offer a more refined estimation. Since this uncertainty type stems from the inherent variability in data, computing it over a subset $\mathcal{S'} \subseteq \mathcal{S}$ rather than individual state-action pairs helps in averaging out the noise and provides a more robust measure:
\begin{equation}
\overline{Var}_A(\hat{M}_{k};\mathcal{S'}) = \frac{1}{\mid \mathcal{S'} \mid} \sum_{(s,a) \in \mathcal{S'}} \left( \frac{1}{N} \sum_{i=1}^{N} \sigma^2_{i}(s,a; W_i) \right)
\end{equation}
where the subscript $A$ refers to aleatoric uncertainty. This method benefits especially in scenarios where individual state-action pairs might have limited observations.

On the other hand, epistemic uncertainty captures the model's lack of knowledge arising from limited data exposure. It stays focused on individual state-action pairs to reveal potentially underrepresented regions in the state-action space:

\begin{equation}
Var_E(\hat{M}_{k};s,a) = \frac{ \sum_{i=1}^{N} \left( \mu_{i}(s,a; W_i) - \overline{\mu}_{k}(s,a) \right)^2}{N-1}
\end{equation}
where the subscript $E$ refers to epistemic uncertainty.

We use exogenous thresholds $\epsilon_E$ and $\epsilon_A$ for epistemic uncertainty and aleatoric uncertainty, respectively, to guide the agent's operational mode. When the differences in these uncertainties between consecutive MDP approximations are below these thresholds, the agent perceives the environment as sufficiently understood and stable to prioritize reward maximization, adopting a regular approach. If the differences exceed these thresholds, indicating heightened uncertainty in the model's understanding or the environment's inherent randomness, the agent adopts a more cautious, worst-case stance. Setting larger thresholds can prompt the agent to shift towards a reward-seeking mode sooner, capitalizing on opportunities but at the potential cost of larger risks.

\subsection{Adaptive Monte Carlo Tree Search}
\begin{algorithm}
\small
\caption{Overall Approach}
\label{algo:train_model}
\textbf{Input:} Global replay buffer \(D\), $\hat{M}_{k}$: model for approximating ${M}_{k}$; $\hat{M}_{k-1}$: model for approximating ${M}_{k-1}$; $N_{\text{interval}}$: model tuning frequency; $N_{\text{threshold}}$: data amount for starting training; $N_u$: training steps
\begin{algorithmic}[1]
\Procedure{Act as you learn}{$D$}
    \State Draw new $w_b \sim P_W$
    \State Init. instance replay buffer $D_b$
    \State $W_{k} \leftarrow W_{k-1}$  \Comment{Transfer weights from $\hat{M}_{k-1}$ to $\hat{M}_{k}$}
    \For{$i = 0$ to $t$ in $M_{k}$}
        \Repeat
            \State $a^* \leftarrow \arg\max_{a \in \mathcal{A}}\text{Ada-MCTS}(s,\hat{M}_{k-1},\hat{M}_{k})$            \State Store $D, D_b \leftarrow (s, a^*, r, s', w_b)$
        \Until{episode is complete}
        \If{$i\mod N_{\text{interval}} = 0 \wedge |D_b| \geq N_{\text{threshold}}$}
        \State \small{$D_b, W_{k}, w_b \leftarrow \text{TuneModel}(D_b, W_{k}, w_b)$}
        \EndIf
    \EndFor
\EndProcedure
\Function{TuneModel}{$D_b, W_{k}, w_b$}
    \For{$k = 0$ to $N_u$ updates}
        \State Update $w_b$ from $D_b$
        \State Update $W_{k}$ from $D_b$
    \EndFor
    \State \Return $D_b, W_{k}, w_b$
\EndFunction
\end{algorithmic}
\end{algorithm}
\begin{algorithm*}
\small
\caption{Adaptive MCTS}
\label{algo:cogmcts}
\textbf{Input:} 
$s_0$: current state; m: total number of simulations; $C_p$: exploration term; $\hat{M}_{k+1}$: model for approximating ${M}_{k+1}$;\\
$\hat{M}_{k}$: model for approximating ${M}_{k}$
\begin{multicols}{2}
\begin{algorithmic}[1]
\Function{Ada-MCTS}{$s_0$,$\hat{M}_{k-1}$,$\hat{M}_{k}$}
    \State create root node $v_0$ with state $s_0$
    \For{$i = 0, 1, \dots, m$}
        \State $v_l \leftarrow \text{Traverse}(v_0)$
        \If{$v_l.s$ is terminal}
            \State $\Delta \leftarrow R(s)$
        \Else
            \State $\Delta \leftarrow \text{Rollout}(v_l.s)$
        \EndIf
        \State \text{Backpropagate}$(v_l, \Delta)$
    \EndFor
    \State return $\pi(a|s_0) \longleftarrow \frac{N(\nu^{s_0,a})}{N(s_0)}$ 
\EndFunction

\Function{Traverse}{$\nu$}
    \While{$\nu$ is nonterminal}
        \If{$\nu$ is DecisionNode}
            \If{$\nu$ is not fully expanded}
                \For{$a \in A$}
                \State $\nu^{\prime} \leftarrow (N_{\text{init}}(s, a), Q_{\text{init}}(s, a))$
                \EndFor    
                \State $\nu^{\prime\prime}.s \leftarrow \text{DPAS}(\nu^{\prime}.s, \nu^{\prime}.a)$
            \Else
            \State $\nu^{\prime} \leftarrow \text{UCT}(\nu, C_p)$
        \EndIf
        \Else
        \State $\nu^{\prime}.s \leftarrow \text{DPAS}(\nu.s, \nu.a)$
        \EndIf
        \State $\nu \leftarrow \nu^{\prime} or \nu^{\prime\prime}$
    \EndWhile
    \State \Return $\nu$
\EndFunction

\columnbreak

\Function{UCT}{$\nu, c$}
    \State \Return $\arg\max_{\nu^{s,a} \in \nu} \left[ V(\nu^{s,a}) + C_p \sqrt{\frac{\log(N(\nu^{s}))}{N(\nu^{s,a})}} \right]$
\EndFunction

\Function{Rollout}{$s$}
    \State $a \sim \mathcal{U}(A)$
    \State $s' \leftarrow \text{DPAS}(s,a)$
    \If{$s'$ is terminal}
        \State \Return $R(s,a)$
    \EndIf
    \State $r \leftarrow R(s,a)$
    \State \Return $r+\gamma\cdot\text{Rollout}(s^{\prime})$
\EndFunction

\Function{Backpropagate}{$\nu, \Delta$}
    \While{$\nu$ is not null}
        \State $N(\nu) \leftarrow N(\nu) + 1$
        \State $V(\nu) \leftarrow V(\nu) + \Delta$
        \State $\nu \leftarrow$ parent of $\nu$
        \State $\Delta \leftarrow \gamma \cdot \Delta$
    \EndWhile
\EndFunction

\Function{DPAS}{$s,a$}
    \State $\delta_E \leftarrow Var_{E}(\hat{M}_{k}(s,a)) - Var_{E}(\hat{M}_{k-1}(s,a))$
    \State $\delta_A \leftarrow \overline{Var}_{A}(\hat{M}_{k};P) - \overline{Var}_{A}(\hat{M}_{k-1};P)$
    \If{$\delta_E \leq \epsilon_E \wedge \delta_A \leq \epsilon_A$}
        \State $s^{\prime} \sim p_(\cdot | s, a,\hat{M}_{k})$
    \Else
        \State $s^{\prime} \sim p_{wc}(\cdot | s, a,\hat{M}_{k-1})$
    \EndIf
    \State \Return $s^{\prime}$
\EndFunction

\end{algorithmic}
\end{multicols}
\end{algorithm*}

Equipped with RA-MCTS and the dual-phase adaptive sampling strategy, a pressing challenge in our setup is the rapid derivation of $\hat{M}_{k}$ that can accurately support decision-making in $M_{k}$. An intuitive approach might involve directly transferring all prior knowledge from $\hat{M}_{k-1}$ to the new model and then collecting new data in $M_{k}$ using RA-MCTS to refine this model. However, such indiscriminate transfers can pose significant complications. Transferring knowledge without accounting for the specific nuances of the evolving dynamics can create a mismatch. This can lead to decisions that, while optimal for $M_{k-1}$, turn out to be suboptimal or even detrimental for $M_{k}$. Moreover, the uncertainties from the previous model would also be carried over, resulting in inaccurate uncertainty estimations for the new model.

To address these challenges, we adopt the latent parameters approach proposed in~\cite{DBLP:conf/aaai/KillianKD17}, as illustrated in Algorithm~\ref{algo:train_model}. The algorithm initiates by drawing a set of new latent parameters $w_b$ from $P_W$, which could be a standard Gaussian distribution. These parameters encapsulate the distinct characteristics of the current environment dynamics. By integrating these parameters into our model, we can quickly pinpoint and adapt to the distinct dynamics of the new environment without forgetting the \textit{general knowledge} from $M_{k-1}$ by transferring the weights from the previous model $\hat{M}_{k-1}$ to the new model $\hat{M}_{k}$. To actuate this idea, we initialize an instance-specific replay buffer $D_b$. As the agent interacts with $M_{k}$, it employs the ADA-MCTS method, detailed in Algorithm~\ref{algo:cogmcts}, which integrates both the prior knowledge from $\hat{M}_{k-1}$ and the current model $\hat{M}_{k}$ for decision-making. The experiences are stored in both the global replay buffer $D$ and the instance-specific replay buffer $D_b$.

By modeling the interaction between the system's state and these latent parameters, our model can effectively generalize the shared dynamics of $\hat{M}_{k-1}$ while adeptly adapting to the distinct dynamics underlying $M_{k}$. The global replay buffer, enriched with data from different instances, enables the model to train on a diverse set of experiences, promoting faster adaptation and scalability. This diversity not only facilitates rapid adaptation to $M_{k}$ but also boosts the model's ability to generalize across various instances and dynamics. In essence, by intertwining the use of latent parameters with structured model adaptation and a diverse global replay buffer, our approach ensures that $\hat{M}_{k}$ is both informed by historical knowledge and finely attuned to the unique dynamics of the current environment, fostering fast and scalable decision-making as the environment evolves.

\section{Experiments}

\begin{table*}[!ht]
\small
\begin{tabular}{@{}l|l|ll|llll@{}}
\multirow{2}{*}{Environment} & \multirow{2}{*}{Setting} & \multicolumn{2}{c|}{\begin{tabular}[c]{@{}c@{}}Approaches that know \\ ground truth transition\end{tabular}} & \multicolumn{4}{c}{Approaches that do not know ground truth transition}                      \\
                             &                          & MCTS-$P_{k}$                                          & RATS-$P_{k}$                                         & MCTS-$\hat{P}_{k-1}$   & RATS-$P_{k-1}$        & RATS-$\hat{P}_{k-1}$ & ADA-MCTS               \\ \midrule
Cliff Walking                & 0.4                      & 0.629 ± 0.11                                          & 0.650 ± 0.04                                         & -0.518 ± 0.17          & 0.079 ± 0.17          & 0.630 ± 0.04       & \textbf{0.778 ± 0.02}  \\
Cliff Walking                & 0.5                      & 0.332 ± 0.18                                          & 0.605 ± 0.03                                         & -0.304 ± 0.20          & 0.384 ± 0.15          & 0.664 ± 0.04       & \textbf{0.815 ± 0.02}  \\
Cliff Walking                & 0.6                      & 0.342 ± 0.18                                          & 0.670 ± 0.04                                         & -0.292 ± 0.20          & 0.253 ± 0.15          & 0.607 ± 0.04       & \textbf{0.830 ± 0.01}  \\
Cliff Walking                & 0.8                      & 0.464 ± 0.18                                          & 0.647 ± 0.08                                         & 0.279 ± 0.20           & 0.625 ± 0.08          & 0.680 ± 0.04       & \textbf{0.871 ± 0.01}  \\
Cliff Walking                & 0.9                      & 0.564 ± 0.17                                          & 0.614 ± 0.07                                         & 0.561 ± 0.17           & 0.758 ± 0.02          & 0.622 ± 0.04       & \textbf{0.883 ± 0.01}  \\
Cliff Walking                & 1.0                      & 0.951 ± 0.00                                          & 0.750 ± 0.04                                         & \textbf{0.947 ± 0.00}           & 0.000 ± 0.00          & 0.000 ± 0.00       & 0.694 ± 0.00  \\
NS Bridge                    & 0.4                      & -0.697 ± 0.02                                         & -0.660 ± 0.03                                        & -0.707 ± 0.03          & -0.684 ± 0.03         & -0.684 ± 0.03      & \textbf{-0.642 ± 0.03} \\
NS Bridge                    & 0.5                      & -0.436 ± 0.13                                         & -0.624 ± 0.03                                        & -0.499 ± 0.12          & -0.590 ± 0.05         & -0.590 ± 0.05      & \textbf{-0.499 ± 0.08} \\
NS Bridge                    & 0.6                      & -0.458 ± 0.13                                         & -0.563 ± 0.04                                        & \textbf{-0.384 ± 0.14} & -0.512 ± 0.05         & -0.512 ± 0.05      & -0.429 ± 0.09          \\
NS Bridge                    & 0.7                      & 0.030 ± 0.16                                          & -0.256 ± 0.05                                        & -0.082 ± 0.16          & -0.206 ± 0.06         & -0.206 ± 0.06      & \textbf{-0.075 ± 0.10} \\
NS Bridge                    & 0.9                      & 0.245 ± 0.15                                          & -0.071 ± 0.04                                        & \textbf{0.135 ± 0.15}  & -0.044 ± 0.03         & -0.044 ± 0.03      & 0.109 ± 0.08           \\
NS Bridge                    & 1.0                      & 0.729 ± 0.00                                          & 0.000 ± 0.00                                         & \textbf{0.729 ± 0.00}  & 0.000 ± 0.00          & 0.000 ± 0.00       & 0.183 ± 0.02           \\
Frozen lake                  & 0.4                      & -0.882 ± 0.10                                         & 0.401 ± 0.16                                         & -0.892 ± 0.10          & 0.264 ± 0.19          & 0.225 ± 0.18       & \textbf{0.426 ± 0.12}  \\
Frozen lake                  & 0.5                      & -0.695 ± 0.16                                         & 0.422 ± 0.16                                         & -0.695 ± 0.16          & 0.239 ± 0.18          & 0.336 ± 0.17       & \textbf{0.446 ± 0.13}  \\
Frozen lake                  & 0.6                      & -0.497 ± 0.19                                         & 0.370 ± 0.15                                         & -0.302 ± 0.21          & \textbf{0.528 ± 0.14} & 0.500 ± 0.13       & 0.474 ± 0.07           \\
Frozen lake                  & 0.8                      & 0.491 ± 0.19                                          & 0.485 ± 0.09                                         & -0.003 ± 0.22          & 0.514 ± 0.06          & 0.416 ± 0.09       & \textbf{0.516 ± 0.11}  \\
Frozen lake                  & 0.9                      & 0.689 ± 0.16                                          & 0.146 ± 0.07                                         & 0.392 ± 0.20           & 0.169 ± 0.05          & 0.169 ± 0.05       & \textbf{0.490 ± 0.12}  \\
Frozen lake                  & 1.0                      & 0.988 ± 0.00                                          & 0.889 ± 0.02                                         & \textbf{0.988 ± 0.00}
           & 0.000 ± 0.00          & 0.000 ± 0.00       & 0.782 ± 0.02  \\ \bottomrule
\end{tabular}
\caption{Expected discounted rewards for all environments using different configurations for MCTS, RATS, and ADA-MCTS. The first two approaches, i.e., MCTS-$P_k$ and RATS-$P_k$ are provided access to current ground truth dynamics, while the others use estimates of the transition dynamics. We observe that ADA-MCTS comprehensively outperforms other approaches. While MCTS-$P_k$ and RATS-$P_k$ are not fair baselines as ADA-MCTS lacks access to ground truth dynamics, we show that ADA-MCTS outperforms RATS in all but one experiment even when RATS has access to ground truth dynamics.}
\label{tab:baseline}
\end{table*}

\begin{figure*}
    \centering
    \includegraphics[width=\textwidth]{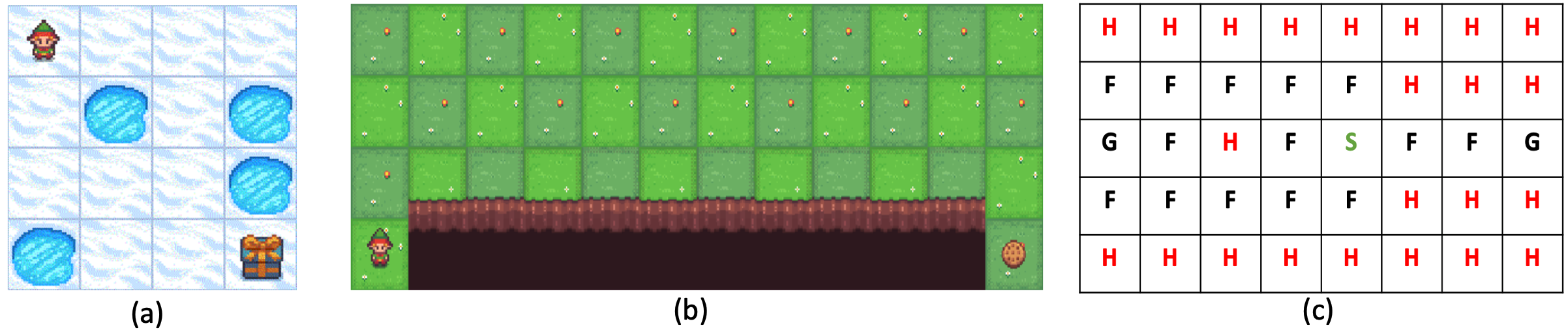}
    \caption{The three environments we use to validate our approach: (a) the frozen lake environment, where the agent must go from start to the goal without falling into the holes; (b) the cliff walking environment, where the agent must go from start to goal without falling off the cliff, but it must accrue a small penalty for each step it takes; and (c) the non-stationary bridge environment from \citet{DBLP:conf/nips/LecarpentierR19}. We add an extra hole to make the environment more challenging. The agent must go from S to G through F (H denotes holes).}
    \label{fig:env}
\end{figure*}

\textbf{Environments:} We use three open-source environments to evaluate our approach (shown in Figure~\ref{fig:env}). We begin by describing the environments below:

\begin{enumerate}[leftmargin=*]
    \item \textbf{Frozen Lake:} The frozen lake environment from Open AI Gym~\citep{brockman2016openai} involves an agent walking from a start position to an end position on a slippery surface. We use a 4x4 frozen lake map, the default environment in Gym. The agent only gets a reward for reaching the goal, and there are no penalties otherwise unless the agent falls into a hole, for which the agent must bear a cost, and the episode ends. At each step, the agent can move either up, down, right, or left. We induce non-stationarity by making the environment more or less slippery.
    \item \textbf{Cliff Walking:} Taken from the Open AI Gym~\citep{brockman2016openai}, this environment involves an agent trying to reach from a start position to the end position without falling into a cliff. At each step, the agent can go right, left, up, or down. The surface can be slippery, preventing the agent from going in its desired direction. Unlike Frozen Lake, the agent concedes a penalty for each step it takes (except the goal, which provides a large positive reward). We induce non-stationarity by making the surface more or less slippery. Note that the gym environment for cliff walking is deterministic; as a result, we manually created the stochastic environment (the environment is available with our code).
    \item \textbf{Non-Stationary Bridge:} For a fair comparison with prior work by \citet{DBLP:conf/nips/LecarpentierR19}, we also evaluate the proposed approach on the non-stationary bridge environment~\cite{DBLP:conf/nips/LecarpentierR19}. In this domain, the agent, in a non-stationary world, faces a situation where no policy is entirely safe.
\end{enumerate}

While these domains are closely related, they present distinct decision-making situations. For example, in the cliff walking domain, the penalty makes the dichotomy between the maximum reward path and the safe path more explicit, unlike the frozen lake environment. Similarly, while the first two environments underscore the challenges of managing exploration versus exploitation and gauging the scalability of algorithms, the bridge environment poses a unique challenge to the worst-case approach by investigating how algorithms behave in settings where no policy is entirely safe. Our implementation is available at \texttt{
https://github.com/scope-lab-vu/ADA-MCTS.git}.

For the frozen lake and the cliff walking environments, the agent's environment can be represented by a probability $p$, which denotes the likelihood that the agent will be able to move in its intended direction; otherwise, it moves in a direction perpendicular to its intended direction with probability $(1-p)/2$. The non-stationary bridge environment~\citep{DBLP:conf/nips/LecarpentierR19} is trickier---the environment can be denoted by a single scalar $p$ with which the agent can move in its intended direction, but it goes in the opposite direction with probability $1-p$.
For all environments, we set the ``original'' environment with $p = 0.7$, and introduce non-stationarity by changing $p$ among $\{0.4,0.5,0.6,0.8,0.9,1\}$.

\textbf{Baselines:} We compare our proposed approach directly with the risk-averse tree search (RATS) algorithm proposed by \citet{DBLP:conf/nips/LecarpentierR19}. RATS uses an \textit{expectminmax} strategy in conjunction with a worst-case approach to determine optimal actions. In order to comprehensively evaluate the proposed approach, we use several scenarios. Consider that the agent is currently in MDP $M_k$ and let $M_{k-1}$ denote the environment immediately before. As before, we use $p_k$ to denote the ground truth transition function for MDP $M_k$, and use $\hat{p}_k$ for the agent's estimate of $p_k$. We evaluate RATS with $p_{k-1}$ (i.e., without adapting to the new environment);\footnote{Note that the approach to adapt to the new environment is, in fact, our contribution.} RATS with $\hat{p}_{k-1}$ (we provide the baseline approach our estimated model from the last MDP); RATS with the current ground truth model (note that this fundamentally handicaps our approach as RATS has access to current ground-truth dynamics); and standard MCTS with $\hat{p}_{k-1}$ (to mimic what an agent would have done had it not had access to an approach for adapting to the new environment).
We also show results with standard MCTS using ${P}_{k}$, i.e., we run regular MCTS with updated ground-truth dynamics. Naturally, this approach is not meant to be a baseline; if the agent has access to ground truth dynamics in the current environment, there is no need for Ada-MCTS. Nonetheless, we show its performance for the sake of comparison. 

\textbf{Hyper-parameters:} 
We conducted all experiments, including MCTS, with 30000 iterations and set the tree depth for RATS at 3 (higher depths were prohibitively expensive computationally, taking hours to compute a single decision). For ADA-MCTS, we configured the parameters as follows: $N_{\text{threshold}}$ is set to 50, $N_{\text{interval}}$ to 5, and $N_u$ to 2. Additionally, we set $\epsilon_{E}$ to 0.02 and $\epsilon_{A}$ to 0 for all the ADA-MCTS experiments.

\begin{figure}[t]
    \centering
    \includegraphics[width=0.65\columnwidth]{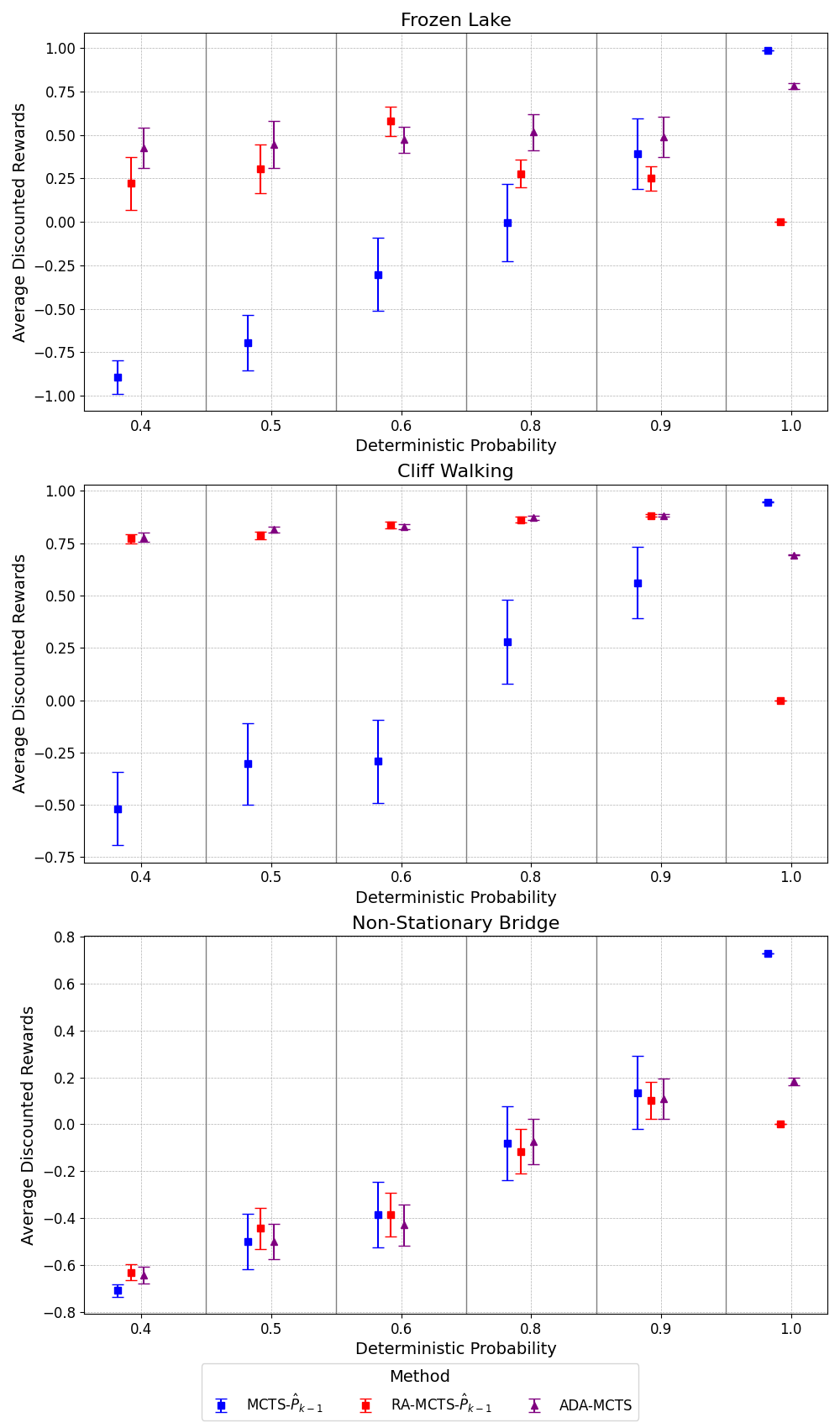}
    \caption{We conduct an ablation study to evaluate the importance of each component of our approach. The experiment highlights the need for effective knowledge transfer and risk-averse exploration.}
    \label{fig:ablation}
\end{figure}

\textbf{Results:}
We show consolidated results in Table~\ref{tab:baseline} (figures for all results are provided in the appendix). In the cliff walking environment, we observe that ADA-MCTS consistently outperforms both RATS and MCTS, regardless of their models or the environment (i.e., $p$.) Incredibly, this dominance persists even when RATS has access to the ground-truth model $p_{k}$, except in a single scenario when $p=1$, i.e., in a purely deterministic environment. We observe the same trend in the non-stationary bridge and the frozen lake environments. \textbf{We reiterate the critical finding---not only does our proposed approach outperform baselines in non-stationary environments, but it also outperforms existing state-of-the-art even when it is grossly handicapped}---while RATS-$P_k$ has access to ground truth environmental dynamics, ADA-MCTS must explore, collect data, and adapt its decision-making to the new environment.
This observed trend can be attributed to the foundational dynamics of ADA-MCTS. At the onset of changing environmental dynamics, ADA-MCTS primarily relies on the base model $\hat{p}_{k-1}$. Hence, before any action, the agent relies on RA-MCTS for data collection. It only switches to regular sampling once it has adapted to the new environment based on its estimates of aleatoric and epistemic uncertainties. 

Notably, RATS, when equipped with $\hat{p}_{k-1}$, and even with $p_{k-1}$, has an average discounted reward of zero in the frozen lake and the cliff walking environments when $p=1$, i.e, the environment becomes fully deterministic. This behavior is explained by the inherent risk-averse strategy of RATS. However, ADA-MCTS achieves high rewards as it slowly collects data using a risk-averse strategy. Once its aleatoric and epistemic uncertainties are within a threshold, it transitions to a ``reward-maximizing'' mode, finding the goal easily in a deterministic environment. Even in environments with more randomness (i.e., $p<0.7$), we observe that the ADA-MCTS can comprehensively outperform RATS and standard MCTS.  We point out one notable exception to the general trend. In the non-stationary bridge environment, we observe that standard MCTS can outperform ADA-MCTS in certain situations. This is a consequence of the fact that no policy is safe in the bridge environment; the ever-present
and unavoidable danger of falling into a hole means that taking a risk-averse approach does not provide a distinct advantage.
Finally, \textbf{we also show that ADA-MCTS is $75\%$ faster than RATS} (average computation time across the three environments are shown in Table~\ref{tab:run}).

\textbf{Ablation Study}:
To evaluate the importance of each component of our proposed approach, we conduct an ablation study. In our ablation study, we evaluate the performance of ADA-MCTS against standard MCTS with $\hat{p}_{k-1}$ (no risk-averse behavior and no knowledge transfer) and RA-MCTS with $\hat{p}_{k-1}$ (risk-averse behavior without knowledge transfer). We show the results in Figure~\ref{fig:ablation}. Notably, in environments like frozen lake and cliff walking with heightened randomness, RA-MCTS significantly outperforms MCTS, a trend similarly observed with ADA-MCTS, highlighting the need for risk-averse decision-making. However, as these environments edged towards determinism, ADA-MCTS began to surpass RA-MCTS, attributed to the agent's transition towards planning using estimated $\hat{p}_k$. In the non-stationary bridge environment, the performance of the three methods converges, especially when $p \neq 1$. This convergence can be understood by considering the inherent structure of the bridge environment; given that no policy guarantees safety and the constant threat of falling into a hole remains, adopting a risk-averse strategy does not necessarily offer a marked benefit, as we discussed before.

\begin{table}[h]
\centering
\small
\begin{tabular}{@{}lll@{}}
\toprule
              & ADA-MCTS & RATS   \\ \midrule
Frozen Lake   & 6.66s    & 23.66s \\
Cliff Walking & 6.33s    & 20.12s \\
NSBridge      & 3.73s    & 24.65s \\ \bottomrule
\end{tabular}
\caption{Average running time for computing a single decision in ADA-MCTS vs. RATS}
\label{tab:run}
\end{table}
\vspace{-0.2cm}

\section{Related Work}

Sequential decision-making in non-stationary environments is well-explored. Early work centered around imposing constraints on the transition probabilities, restricting them to predefined polytopes~\citep{satia1973markovian, white1994markov}.  The crucial question of how to construct these polytopes remained unanswered, as underscored by \citet{iyengar2005robust}, who introduced the concept of robust Markov Decision Processes (MDPs), incorporating uncertain priors, wherein the transition function can vary among a set of functions due to inherent uncertainty. Subsequently, \citet{DBLP:conf/nips/LecarpentierR19} proposed a principled model for non-stationary Markov decision processes (NSMDP), extending the scope of prior work by permitting uncertainty in both the reward model and the transition function; in their work, the rate of change was bounded by Lipschitz Continuity. A closely related problem setting is Lifelong Reinforcement Learning (LRL)~\citep{thrun1995lifelong}, in which an agent focuses on learning multiple tasks sequentially, leveraging previous tasks' knowledge to improve the learning of new tasks. While we deal with a similar problem setting, we explore a fundamentally different challenge, i.e., how can an agent use online planning to explore a new environment and adapt to it safely? While \citet{DBLP:conf/nips/LecarpentierR19} propose risk-averse tree search to tackle this challenge, they consider that the environment changes continuously and that the current environment is known to the agent. We look at the discrete environmental changes and explore settings where the agent must learn the transition dynamics by interacting with the environment.
\section{Conclusion}

We present a novel heuristic search strategy based on Monte Carlo tree search that can adapt to non-stationary environments. Our approach uses a risk-averse strategy to explore the new environment safely and uses epistemic and aleatoric uncertainties to switch to reward-maximizing behavior. Through extensive experiments using multiple open-source environments, we observe that our approach outperforms the existing state-of-the-art, even when the latter has access to ground truth dynamics of the environment, which our proposed approach must slowly learn while minimizing risk.

\section*{Acknowledgments}
This material is based upon work sponsored by the National Science Foundation (NSF) under Grant CNS-2238815 and by the Defense Advanced Research Projects Agency (DARPA). Results presented in this paper were obtained using the Chameleon testbed supported by the National Science Foundation. Any opinions, findings, and conclusions or recommendations expressed in this material are those of the authors and do not necessarily reflect the views of the NSF, or the DARPA.

\bibliographystyle{plainnat}
\bibliography{main}

\end{document}